\documentclass[10pt,twocolumn,letterpaper]{article}

\usepackage{cvpr}
\usepackage{times}
\usepackage{epsfig}
\usepackage{graphicx}

\usepackage{color,xcolor}

\usepackage{adjustbox}
\usepackage{array}
\usepackage{booktabs}
\usepackage{colortbl}
\usepackage{float,wrapfig}
\usepackage{hhline}
\usepackage{multirow}

\usepackage{amsmath,amsfonts,amssymb}
\usepackage{bm}
\usepackage{nicefrac}
\usepackage{microtype}

\usepackage{changepage}
\usepackage{extramarks}
\usepackage{fancyhdr}
\usepackage{lastpage}
\usepackage{setspace}
\usepackage{soul}
\usepackage{xspace}

\usepackage{url}

\usepackage{enumerate}
\usepackage[pagebackref=true,breaklinks=true,colorlinks,bookmarks=false]{hyperref}

\newcommand{\supp}{supplemental material\xspace}

\newcommand{\reffig}[1]{Figure~\ref{fig:#1}}
\newcommand{\refsec}[1]{Section~\ref{sec:#1}}

\newcommand{\reftbl}[1]{Table~\ref{tbl:#1}}

\newcommand{\refeq}[1]{Equation~\ref{eq:#1}}

\newcommand{\lblfig}[1]{\label{fig:#1}}
\newcommand{\lblsec}[1]{\label{sec:#1}}
\newcommand{\lbleq}[1]{\label{eq:#1}}
\newcommand{\lbltbl}[1]{\label{tbl:#1}}

\newcommand{\ignorethis}[1]{}
\newcommand{\pp}{pix2pix\xspace}
\newcommand{\G}{G\xspace}
\newcommand{\D}{D\xspace}
\newcommand{\E}{\mathbb{E}\xspace}

\newcommand{\x}{\mathbf{x}}
\newcommand{\y}{\mathbf{y}}
\newcommand{\rrr}{\mathbf{r}}
\newcommand{\www}{\mathbf{w}}
\newcommand{\myparagraph}[1]{\vspace{-12pt}\paragraph{#1}}

\cvprfinalcopy 

\def\cvprPaperID{4903} 

\ifcvprfinal\fi
\begin{document}

\title{Connecting Touch and Vision via Cross-Modal Prediction}

\author{Yunzhu Li \quad Jun-Yan Zhu \quad Russ Tedrake \quad Antonio Torralba\\
MIT CSAIL 
}

\maketitle

\begin{abstract}
Humans perceive the world using multi-modal sensory inputs such as vision, audition, and touch. In this work, we investigate the cross-modal connection between vision and touch.
The main challenge in this cross-domain modeling task lies in the significant scale discrepancy between the two: while our eyes perceive an entire visual scene at once, humans can only feel a small region of an object at any given moment. To connect vision and touch, we introduce new tasks of synthesizing plausible tactile signals from visual inputs as well as imagining how we interact with objects given tactile data as input. To accomplish our goals, we first equip robots with both visual and tactile sensors and collect a large-scale dataset of corresponding vision and tactile image sequences. To close the scale gap, we present a new conditional adversarial model that incorporates the scale and location information of the touch. Human perceptual studies demonstrate that our model can produce realistic visual images from tactile data and vice versa. Finally, we present both qualitative and quantitative experimental results regarding different system designs, as well as visualizing the learned representations of our model.

\vspace{-10pt}

\end{abstract}

\section{Introduction}
\lblsec{intro}
\vspace{-2mm}

\begin{figure}
    \centering
    \includegraphics[width=0.475\textwidth]{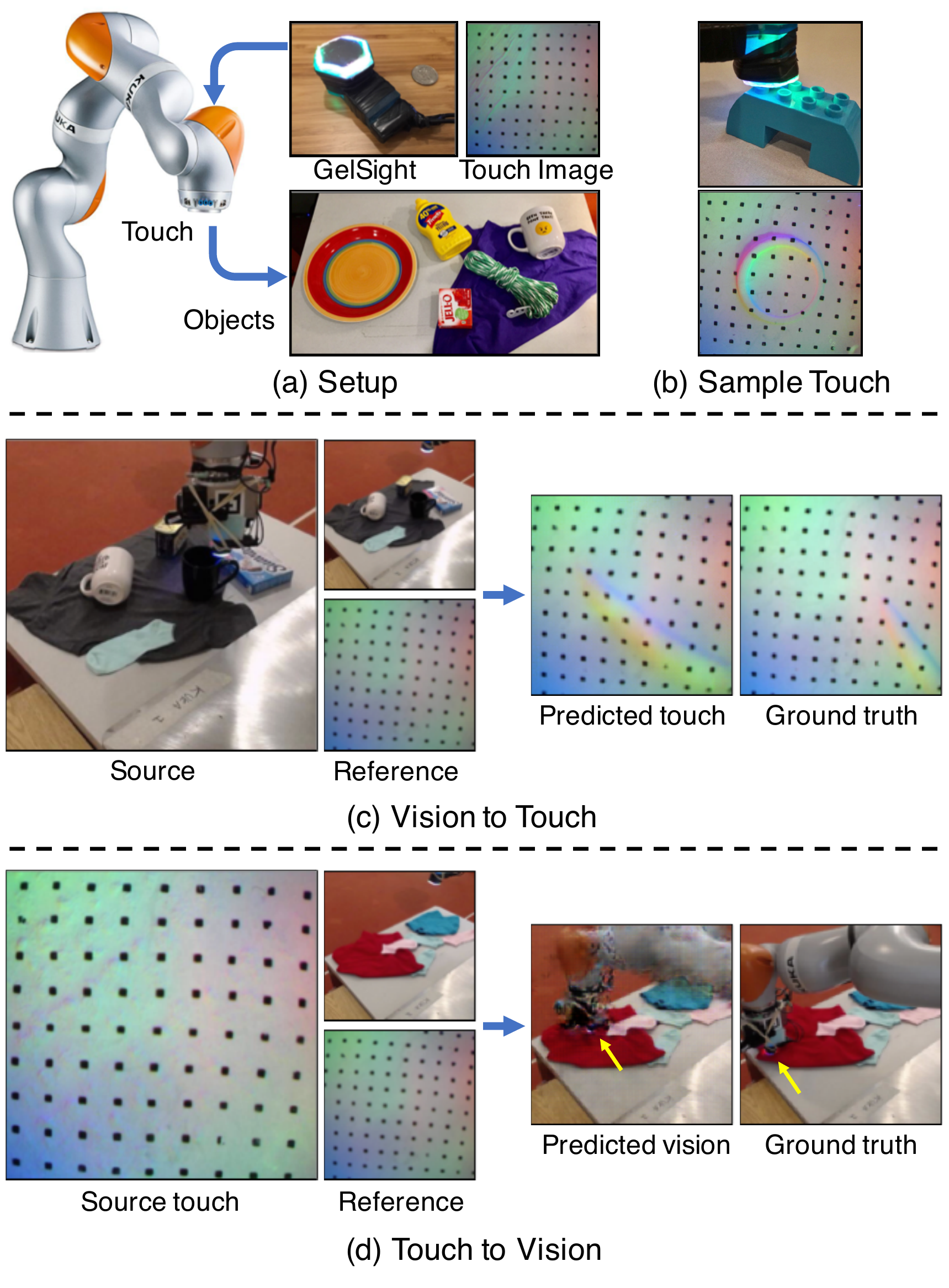}
    \vspace{-15pt}
    \caption{ \small {\bf Data collection setup:} (a) we use a robot arm equipped with a GelSight sensor~\cite{johnson2011microgeometry} to collect tactile data and use a webcam to capture the videos of object interaction scenes. (b) An illustration of the GelSight touching an object. {\bf Cross-modal prediction:} given the collected vision-tactile pairs, we train cross-modal prediction networks for several tasks: (c) Learning to feel by seeing (vision $\rightarrow$ touch): predicting  the a touch signal from its corresponding vision input and reference images and (d) Learning to see by touching (touch $\rightarrow$ vision): predicting vision from touch. The predicted touch locations and ground truth locations (marked with yellow arrows in (d)) share a similar feeling. Please check out our \href{http://visgel.csail.mit.edu/}{website} for code and more results.}
    \lblfig{overview}
    \vspace{-10pt}
\end{figure}

People perceive the world in a multi-modal way where vision and touch are highly intertwined~\cite{lederman1986perception,yau2009analogous}: when we close our eyes and use only our fingertips to sense an object in front of us, we can make guesses about its texture and geometry. For example in \reffig{overview}d, one can probably tell that s/he is touching a piece of delicate fabric based on its tactile ``feeling"; similarly, we can imagine the feeling of touch by just seeing the
object. In \reffig{overview}c, without directly contacting the rim of a mug, we can easily imagine the sharpness and hardness of the touch merely by our visual perception. The underlying reason for this cross-modal connection is the shared physical properties that influence both modalities such as local geometry, texture, roughness, hardness and so on. Therefore, 
it would be desirable to build a computational model that can extract such shared representations from one modality and transfer them to the other.

In this work, we present a cross-modal prediction system between vision and touch with the goals of \textit{learning to see by touching} and \textit{learning to feel by seeing}. Different from other cross-modal prediction problems where sensory data in different domains are roughly spatially aligned~\cite{isola2017image,aytar2017cross}, the scale gap between vision and touch signals is huge. While our visual perception system processes the entire scene as a whole, our fingers can only sense a tiny fraction of the object at any given moment. 
To investigate the connections between vision and touch, we introduce two cross-modal prediction tasks: (1) synthesizing plausible temporal tactile signals from vision inputs, and (2) predicting which object and which object part is being touched directly from tactile inputs. \reffig{overview}c and d show a few representative results.

To accomplish these tasks, we build a robotic system to automate the process of collecting large-scale visual-touch pairs. As shown in \reffig{overview}a, a robot arm is equipped with a tactile sensor called GelSight~\cite{johnson2011microgeometry}. We also set up a standalone web camera  to record visual information of both objects and arms. In total, we recorded $12,000$ touches on $195$ objects from a wide range of categories. Each touch action contains a video sequence of $250$ frames, resulting in $3$ million visual and tactile paired images. The usage of the dataset is not limited to the above two applications.

Our model is built on conditional adversarial networks~\cite{goodfellow2014generative,isola2017image}. The standard approach~\cite{isola2017image} yields less satisfactory results in our tasks due to the following two challenges. First, the scale gap between vision and touch makes the previous methods~\cite{isola2017image,wang2017high} less suitable as they are tailored for spatially aligned image pairs.
To address this scale gap, we incorporate the scale and location information of the touch into our model, which significantly improves the results. Second, we encounter severe mode collapse during GANs training when the model generates the same output regardless of inputs. It is because the majority of our tactile data only contain flat regions as often times, robots arms are either in the air or touching textureless surface, To prevent mode collapse, we adopt a data rebalancing strategy to help the generator produce diverse modes. 

We present both qualitative results and quantitative analysis to evaluate our model. The evaluations include human perceptual studies regarding the photorealism of the results, as well as objective measures such as the accuracy of touch locations and the amount of deformation in the GelSight images. We also perform ablation studies regarding alternative model choices and objective functions. Finally, we visualize the learned representations of our model to help understand what it has captured.
\section{Related Work}
\lblsec{related}
\vspace{10pt}
\myparagraph{Cross-modal learning and prediction}
People understand our visual world through many different modalities. Inspired by this, many researchers proposed to learn shared embeddings from multiple domains such as words and images~\cite{frome2013devise}, audio and videos~\cite{ngiam2011multimodal,aytar2016soundnet,owens2016ambient}, and texts and visual data~\cite{norouzi2013zero,otani2016learning,aytar2017cross}. 
Our work is mostly related to cross-modal prediction, which aims to predict data in one domain from another. Recent work has tackle different prediction tasks such as using vision to predict sound~\cite{owens2016visually} and generating captions for images~\cite{kulkarni2013babytalk,karpathy2015deep,xu2015show,donahue2015long},  thanks to large-scale paired cross-domain datasets, which are not currently available for vision and touch. We circumvent this difficulty by automating the data collection process with robots.

\myparagraph{Vision and touch}
To give intelligent robots the same tactile sensing ability, different types of force, haptic, and tactile sensors~\cite{lederman1987hand,lederman2009haptic,cutkosky2008force,kappassov2015tactile,SSundaram:2019:STAG} have been developed over the decades. Among them, GelSight~\cite{johnson2009retrographic,johnson2011microgeometry,yuan2017gelsight} is considered among the best high-resolution tactile sensors.
Recently, researchers have used GelSight and other types of force and tactile sensors for many vision and robotic applications~\cite{yuan2016estimating,yuan2017shape, yuan2015measurement,li2013sensing,li2014localization,lee2018making}. Yuan et al.~\cite{yuan2017connecting} studied physical and material properties of fabrics by fusing visual, depth, and tactile sensors. Calandra et al.~\cite{calandra2017feeling} proposed a visual-tactile model for predicting grasp outcomes. Different from prior work that used vision and touch to improve individual tasks, in this work we focus on several cross-modal prediction tasks, investigating whether we can predict one signal from the other.

\myparagraph{Image-to-image translation}
Our model is built upon recent work on image-to-image translation~\cite{isola2017image,liu2016unsupervised,zhu2017unpaired}, which aims to translate an input image from one domain to a photorealistic output in the target domain. The key to its success relies on adversarial training~\cite{goodfellow2014generative,mirza2014conditional}, where a discriminator is trained to distinguish between the generated results and real images from the target domain. This method enables many applications such as synthesizing photos from user sketches~\cite{isola2017image,sangkloy2016scribbler}, changing night to day~\cite{isola2017image,zhu2017multimodal}, and turning semantic layouts into natural scenes~\cite{isola2017image,wang2017high}. Prior work often assumes that input and output images are geometrically aligned, which does not hold in our tasks due to the dramatic scale difference between two modalities. Therefore, we design objective functions and architectures to sidestep this scale mismatch.  In \refsec{experiments}, we show that we can obtain more visually appealing results compared to recent methods~\cite{isola2017image}. 

\section{VisGel Dataset}
\lblsec{data}
\vspace{-2mm}
Here we describe our data collection procedure including the tactile sensor we used, the way that robotic arms interacting with objects, and a diverse object set that includes $195$ different everyday items from a wide range of categories. 

\myparagraph{Data collection setup}

\reffig{overview}a illustrates the setup in our experiments. We use KUKA LBR iiwa industrial robotic arms to automate the data collection process.
The arm is equipped with a GelSight sensor~\cite{yuan2017gelsight} to collect raw tactile images. We set up a webcam on a tripod at the back of the arm to capture videos of the scenes where the robotic arm touching the objects. We use recorded timestamps to synchronize visual and tactile images.

\myparagraph{GelSight sensor}

The GelSight sensor~\cite{johnson2009retrographic,johnson2011microgeometry,yuan2017gelsight} is an optical tactile sensor that measures the texture and geometry of a contact surface at very high spatial resolution~\cite{johnson2011microgeometry}. The surface of the sensor is a soft elastomer painted with a reflective membrane that deforms to the shape of the object upon contact, and the sensing area is about $1.5\text{cm} \times 1.5\text{cm}$. Underneath this elastomer is an ordinary camera that views the deformed gel. The colored LEDs illuminate the gel from different directions, resulting in a three-channel surface normal image (\reffig{overview}b). GelSight also uses markers on the membrane and recording the flow field of the marker movement to
sketched the deformation. The 2D image format of the raw tactile data allows us to use standard convolutional neural networks (CNN)~\cite{lecun1998gradient} for processing and extracting tactile information. \reffig{overview}c and d show a few examples of collected raw tactile data. 

\myparagraph{Objects dataset} \reffig{objects} shows all the $195$ objects used in our study. To collect such a diverse set of objects, we start from Yale-CMU-Berkeley (YCB) dataset~\cite{calli2017ycb}, a standard daily life object dataset widely used in robotic manipulation research. We use $45$ objects with a wide range of shapes, textures, weight, sizes, and rigidity. We discard the rest of the $25$ small objects (e.g., plastic nut) as they are occluded by the robot arm from the camera viewpoint. To further increase the diversity of objects, we obtain additional $150$ new consumer products that include the categories in the YCB dataset (i.e., food items, tool items, shape items, task items, and kitchen items) as well as new categories such as fabrics and stationery. We use $165$ objects during our training and $30$ seen and $30$ novel objects during test. Each scene contains $4\sim 10$ randomly placed objects that sometimes overlap with each other.

\begin{figure}[t]
\centering
\includegraphics[width=0.48\textwidth]{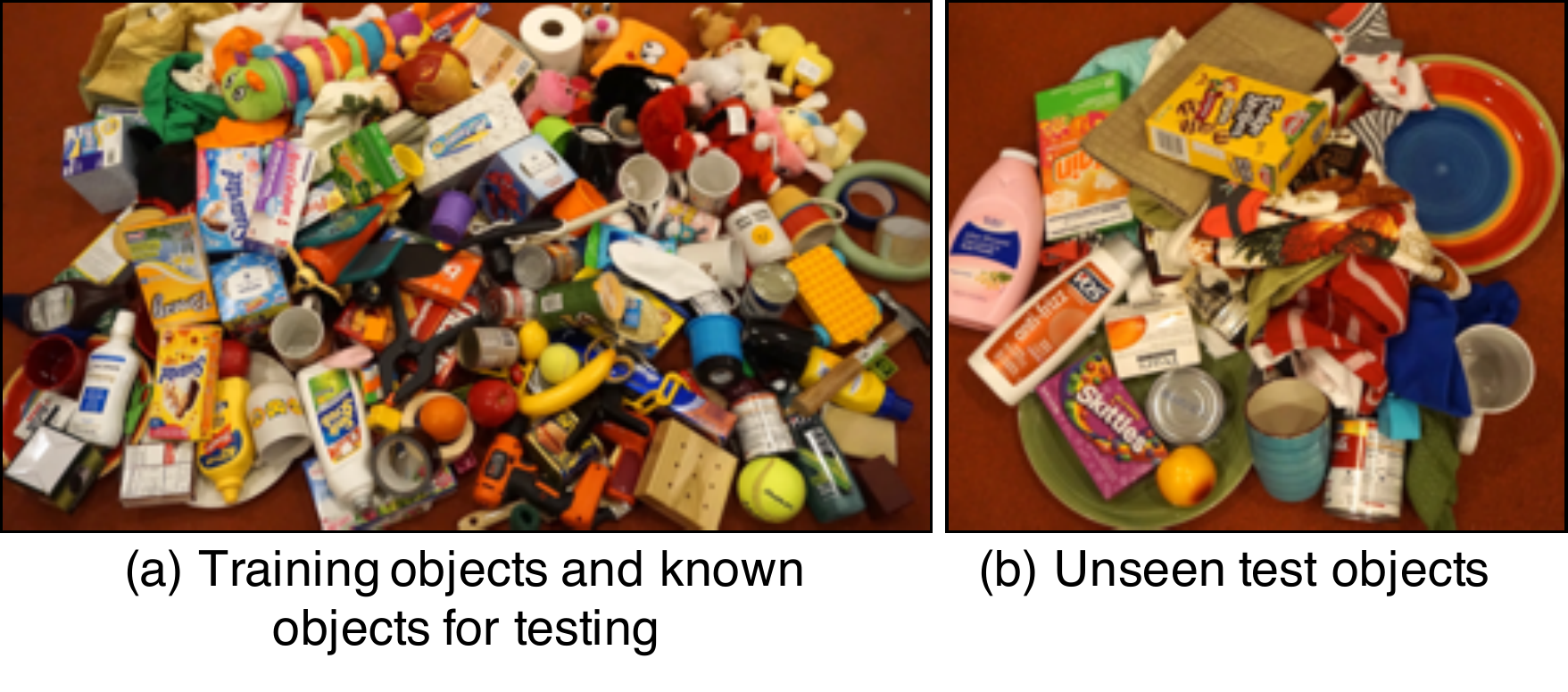}
\vspace{-15pt}
\caption{\small {\bf Object set.} Here we show the object set used in training and test. The dataset includes a wide range of objects from food items, tools, kitchen items, to fabrics and stationery.}
\vspace{-10pt}
\lblfig{objects}
\end{figure}
\myparagraph{Generating touch proposals}
A random touch at an arbitrary location may be suboptimal due to two reasons. First, the robotic arm can often touch nothing but the desk. Second, the arm may touch in an undesirable direction or unexpectedly move the object so that the GelSight Sensor fails to capture any tactile signal. To address the above issues and generate better touch proposals, we first reconstruct 3D point clouds of the scene with a real-time SLAM system called ElasticFusion~\cite{whelan2015elasticfusion}.
We then sample a random touch region whose surface normals are mostly perpendicular to the desk. The touching direction is important as it allows robot arms to firmly press the object without moving it.
\myparagraph{Dataset Statistics}

We have collected synchronized tactile images and RGB images for $195$ objects. \reftbl{dataset} shows the basic statistics of the dataset for both training and test. To our knowledge, this is the largest vision-touch dataset.

\begin{table}[t]
\begin{center}
  \begin{tabular}{ | l || r | r | }
    \hline
    & \# touches & \# total vision-touch frames  \\ \hline
    Train & 10,000 & 2,500,000 \\ \hline
    Test & 2,000 & 500,000 \\
    \hline
  \end{tabular}
\end{center}
\vspace{-5pt}
 \caption{{\bf Statistics of our VisGel dataset.} We use a video camera and a tactile sensor to collect a large-scale synchronized videos of a robot arm interacting with household objects. }
    \lbltbl{dataset}
    \vspace{-10pt}
\end{table}

\section{Cross-Modal Prediction}
\vspace{-2mm}

\begin{figure*}[t]
\centering
\includegraphics[width=\textwidth]{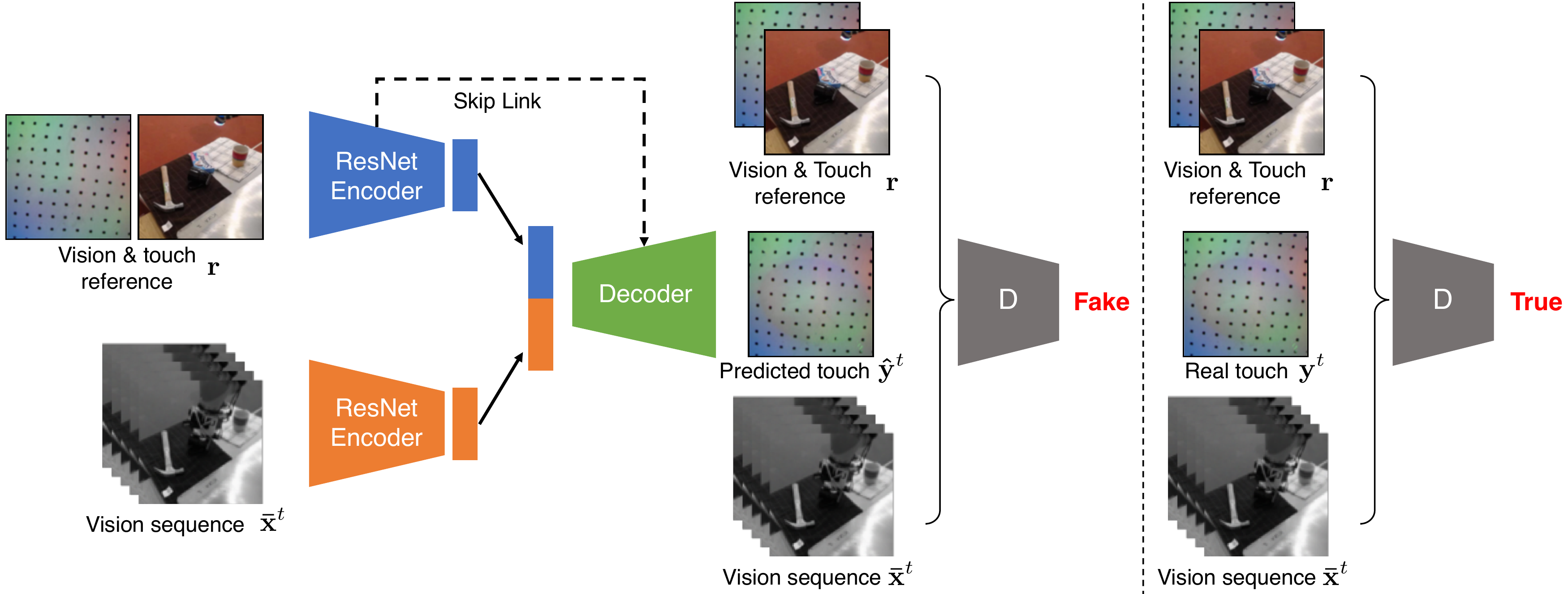}
\vspace{-15pt}

\caption{\small {\bf Overview of our cross-modal prediction model.}
Here we show our vision $\rightarrow$ touch model. The generator $\G$ consists of two ResNet encoders and one decoder. It takes both reference vision and touch images $\mathbf{r}$ as well as a sequence of frames $\bar{\x}^t$ as input, and predict the tactile signal $\hat{\y}^t$ as output. Both reference images and temporal information help improve the results. 
Our discriminator learns to distinguish between the generated tactile signal $\hat{\y}^t$ and real tactile data $\y^t$.  
For touch $\rightarrow$ vision, we switch the input and output modality and train the model under the same framework.}
\lblfig{arch}
\vspace{-10pt}
\end{figure*}
\lblsec{method}

We propose a cross-modal prediction method for predicting vision from touch and vice versa. 
First, we describe our basic method based on conditional GANs~\cite{isola2017image} in \refsec{method:cGANs}. We further improve the accuracy and the quality of our prediction results with three modifications tailored for our tasks in \refsec{method:improve}. We first incorporate the scale and location of the touch into our model. Then, we use a data rebalancing mechanism to increase the diversity of our results. Finally, we further improve the temporal coherence and accuracy of our results by extracting temporal information from nearby input frames. In \refsec{method:details}, we describe the details of our training procedure as well as network designs.

\subsection{Conditional GANs}
\lblsec{method:cGANs}

Our approach is built on the \pp method~\cite{isola2017image}, a recently proposed general-purpose conditional GANs framework for image-to-image translation. 
In the context of vision-touch cross-modal prediction, the generator $\G$ takes either a vision or tactile image $x$ as an input and produce an output image in the other domain with $y=\G(x)$. 
 The discriminator $\D$ observes both the input image $x$ and the output result $y$: $\D(x, y) \rightarrow [0, 1]$. During training, the discriminator $\D$ is trained to reveal the differences between synthesized results and real images while the objective of the generator $\G$ is to produce photorealistic results that can fool the discriminator $\D$. We train the model with vision-touch image pairs $\{(\mathbf{x},\mathbf{y})\}$. In the task of touch $\rightarrow$ vision, $\mathbf{x}$ is a touch image and $\mathbf{y}$ is the corresponding visual image. The same thing applies to the vision $\rightarrow$ touch direction, i.e., $(\mathbf{x},\mathbf{y}) = (\text{visual image}, \text{touch image})$. Conditional GANs can be optimized via the following min-max objective: 
\begin{equation}
\G^*=\arg \min_{\G} \max_{\D} \mathcal{L}_{\text{GAN}}(\G, \D) + \lambda \mathcal{L}_1(\G)
\lbleq{minimax}
\end{equation}
where the adversarial loss $\mathcal{L}_{\text{GAN}}(\G,\D)$ is derived as:
\begin{equation}
\E_{(\x,\y)}[\log \D(\x,\y)]+  \E_{\x}[\log (1- \D(\x,\G(\x))],
\lbleq{gan_problem}
\end{equation}
where the generator $\G$ strives to minimize the above objective against the discriminator's effort to maximize it, and we denote $\mathbb{E}_{\x} \triangleq \mathbb{E}_{\x\sim p_{\text{data}}(x)}$ and $ \mathbb{E}_{(\x,\y)} \triangleq \mathbb{E}_{(\x,\y) \sim p_{\text{data}}(\x, \y)}$ for brevity. In additional to the GAN loss, we also add a direct regression $L_1$ loss between the predicted results and the ground truth images. This loss has been shown to help stabilize GAN training in prior work~\cite{isola2017image}:
\begin{equation}
\mathcal{L}_1(\G) = \E_{(\x,\y)} ||\y - \G(\x)||_1
\lbleq{l1}
\end{equation}

\subsection{Improving Photorealism and Accuracy}
\lblsec{method:improve}

We first experimented with the above conditional GANs framework. Unfortunately, as shown in the \reffig{reference}, the synthesized results are far from satisfactory, often looking unrealistic and suffering from severe visual artifacts. Besides, the generated results do not align well with input signals. 

To address the above issues, we make a few modifications to the basic algorithm, which significantly improve the quality of the results as well as the match between input-output pairs. We first feed tactile and visual reference images to both the generator and the discriminator so that the model only needs to learn to model cross-modal changes rather than the entire signal. Second, we use a data-driven data rebalancing mechanism in our training so that the network is more robust to mode collapse problem where the data is highly imbalanced. Finally, we extract information from multiple neighbor frames of input videos rather than the current frame alone, producing temporal coherent outputs. 

\myparagraph{Using reference tactile and visual images}
As we have mentioned before, the scale between a touch signal and a visual image is huge as a GelSight sensor can only contact a very tiny portion compared to the visual image. This gap makes the cross-modal prediction between vision and touch quite challenging. Regarding touch to vision, we need to solve an almost impossible `extrapolation' problem from a tiny patch to an entire image. From vision to touch, the model has to first localize the location of the touch and then infer the material and geometry of the touched region. \reffig{reference} shows a few results produced by conditional GANs model described in \refsec{method:cGANs}, where no reference is used. The low quality of the results is not surprising due to self-occlusion and big scale discrepancy.

We sidestep this difficulty by providing our system both the reference tactile and visual images as shown in \reffig{overview}c and d. 
A reference visual image captures the original scene without any robot-object interaction. 
For vision to touch direction, when the robot arm is operating, our model can simply compare the current frame with its reference image and easily identify the location and the scale of the touch. For touch to vision direction, a reference visual image can tell our model the original scene and our model only needs to predict the location of the touch and hallucinate the robotic arm, without rendering the entire scene from scratch. 
A reference tactile image captures the tactile response when the sensor touches nothing, which can help the system calibrate the tactile input, as different GelSight sensors have different lighting distribution and black dot patterns.

\begin{figure}[t]
\centering
\includegraphics[width=.48\textwidth]{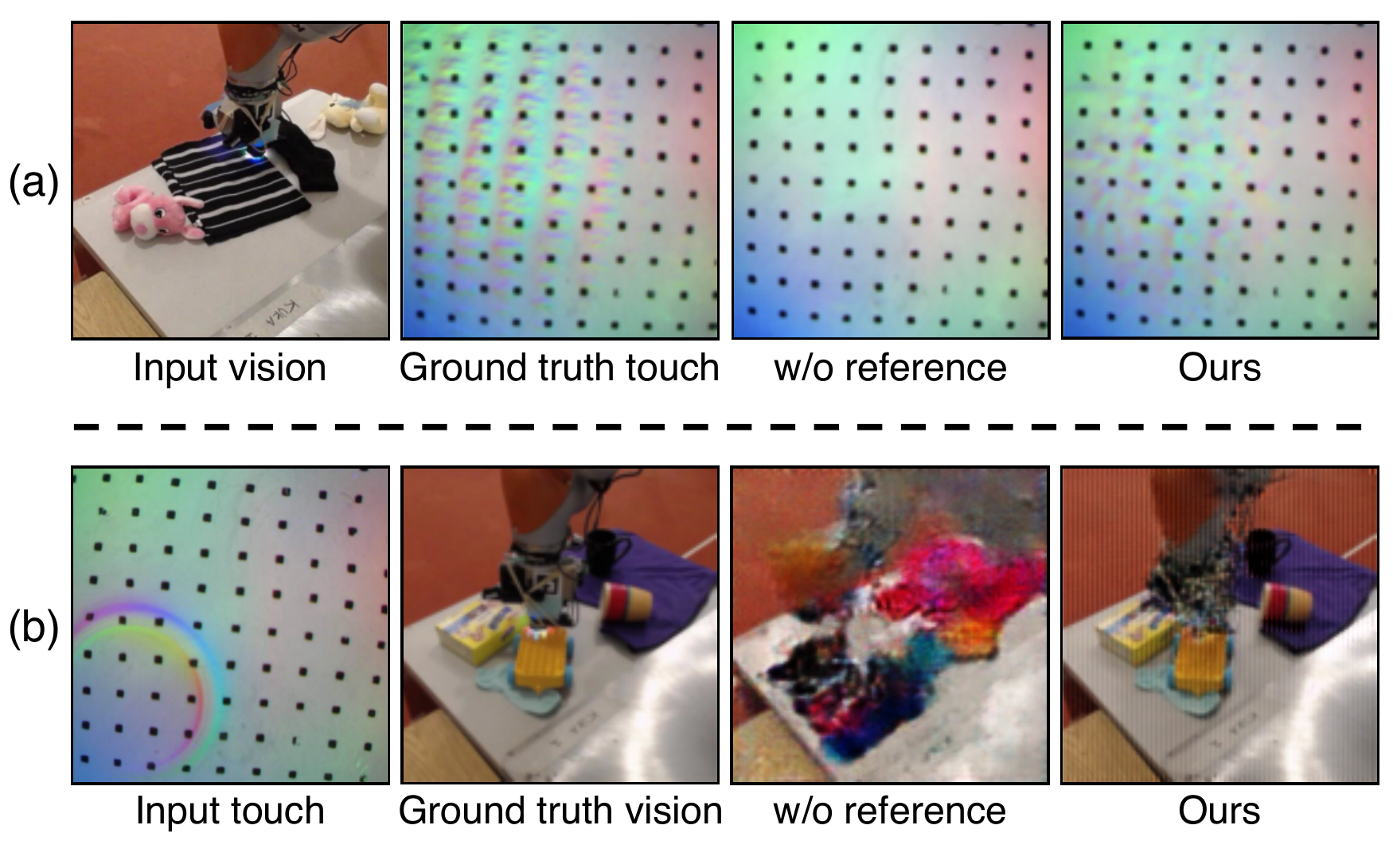}
\vspace{-10pt}
\caption{\small {\bf Using reference images.} Qualitative results of our methods with / without using reference images. Our model trained with reference images produces more visually appealing images.}
\lblfig{reference}
\vspace{-10pt}
\end{figure}

In particular, we feed both vision and tactile reference images  $\rrr = (\x^{\text{ref}}, \y^{\text{ref}})$ to the generator $\G$ and the discriminator $\D$.  As the reference image and the output often share common low-level features, we introduce skip connections~\cite{ronneberger2015u,he2016deep} between the encoder convolutional layers and transposed-convolutional layers in our decoder.

\myparagraph{Data rebalancing}
In our recorded data, around $60$ percentage of times, the robot arm is in the air without touching any object. 
This results in a huge data imbalance issue, where more than half of our tactile data has only near-flat responses without any texture or geometry. This highly imbalanced dataset causes severe model collapse during GANs training~\cite{goodfellow2016nips}. To address it, we apply data rebalancing technique, widely used in classification tasks~\cite{farabet2013learning,zhang2016colorful}. In particular, during the training, we reweight the loss of each data pair $(\x^t, \rrr, \y^t)$ based on its rarity score $\www^t$. In practice, we calculate the rarity score based on a ad-hoc metric. We first compute a residual image $\|\x^t - \x^{\text{ref}}\|$ between the current tactile data $\x^t$ and its reference tactile data $\x^{\text{ref}}$. We then simply calculate the variance of Laplacian derivatives over the difference image. For IO efficiency, instead of reweighting, we sample the training data pair $(\x^t, \rrr, \y^t)$  with the probability $\frac{\www^t}{\sum_{t} \www^t}$. We denote the resulting data distribution as $p_{\www}$. \reffig{quali} shows a few qualitative results  demonstrating the improvement by using data rebalancing. Our evaluation in \refsec{experiments} also shows the effectiveness of data rebalancing. 
\myparagraph{Incorporating temporal cues}
We find that our initial results look quite realistic, but the predicted output sequences and input sequences are often out of sync (\reffig{motion_scale}). To address this temporal mismatch issue, we use multiple nearby frames of the input signal in addition to its current frame. In practice, we sample $5$ consecutive frames every $2$ frames: $\bar{\x}^{t} = \{\x^{t-4}, \x^{t-2}, \x^{t}, \x^{t+2}, \x^{t+4}\}$ at a particular moment $t$. To reduce data redundancy, we only use grayscale images and leave the reference image as RGB.  
\myparagraph{Our full model}
\reffig{arch} shows an overview of our final cross-modal prediction model. The generator $\G$ takes both input data $\bar{\x}^{t} = \{\x^{t-4}, \x^{t-2}, \x^{t}, \x^{t+2}, \x^{t+4}\}$ as well as reference vision and tactile images $\rrr = (\x^{\text{ref}}, \y^{\text{ref}})$ and produce a output image $\hat{\y}^t = \G(\bar{\x}^t, \rrr)$ at moment $t$ in the target domain.  We extend the minimax objective (\refeq{minimax}) $\mathcal{L}_{\text{GAN}}(\G, \D) + \lambda \mathcal{L}_1(\G)$, where $\mathcal{L}_{\text{GAN}}(\G, \D)$ is as follows:
\begin{equation}
\E_{(\bar{\x}^t,\rrr, \y^t)\sim p_{\www}
}[\log \D(\bar{\x}^t, \rrr,\y^t)]+
\E_{(\bar{\x}^t,\rrr)\sim p_{\www}
}[\log (1- \D(\bar{\x}^t, \rrr, \hat{\y}^t)],
\lbleq{gan2}
\end{equation}
where $\G$ and $\D$ both takes both temporal data $\bar{\x}^t$ and reference images $\rrr$ as inputs. Similarly, the regression loss $ \mathcal{L}_1(\G)$ can be calculated as:  
\begin{equation}
\mathcal{L}_1(\G) = \E_{(\bar{\x}^t,\rrr,\y^t) \sim p_{\www}} ||\y^t - \hat{\y}^t||_1
\lbleq{l1_2}
\end{equation}

\reffig{arch} shows a sample input-output combination where the network takes a sequence of visual images and the corresponding references as inputs, synthesizing a tactile prediction as output. The same framework can be applied to the touch $\rightarrow$ vision direction as well. 

\begin{figure*}[t]
\centering
\includegraphics[width=\textwidth]{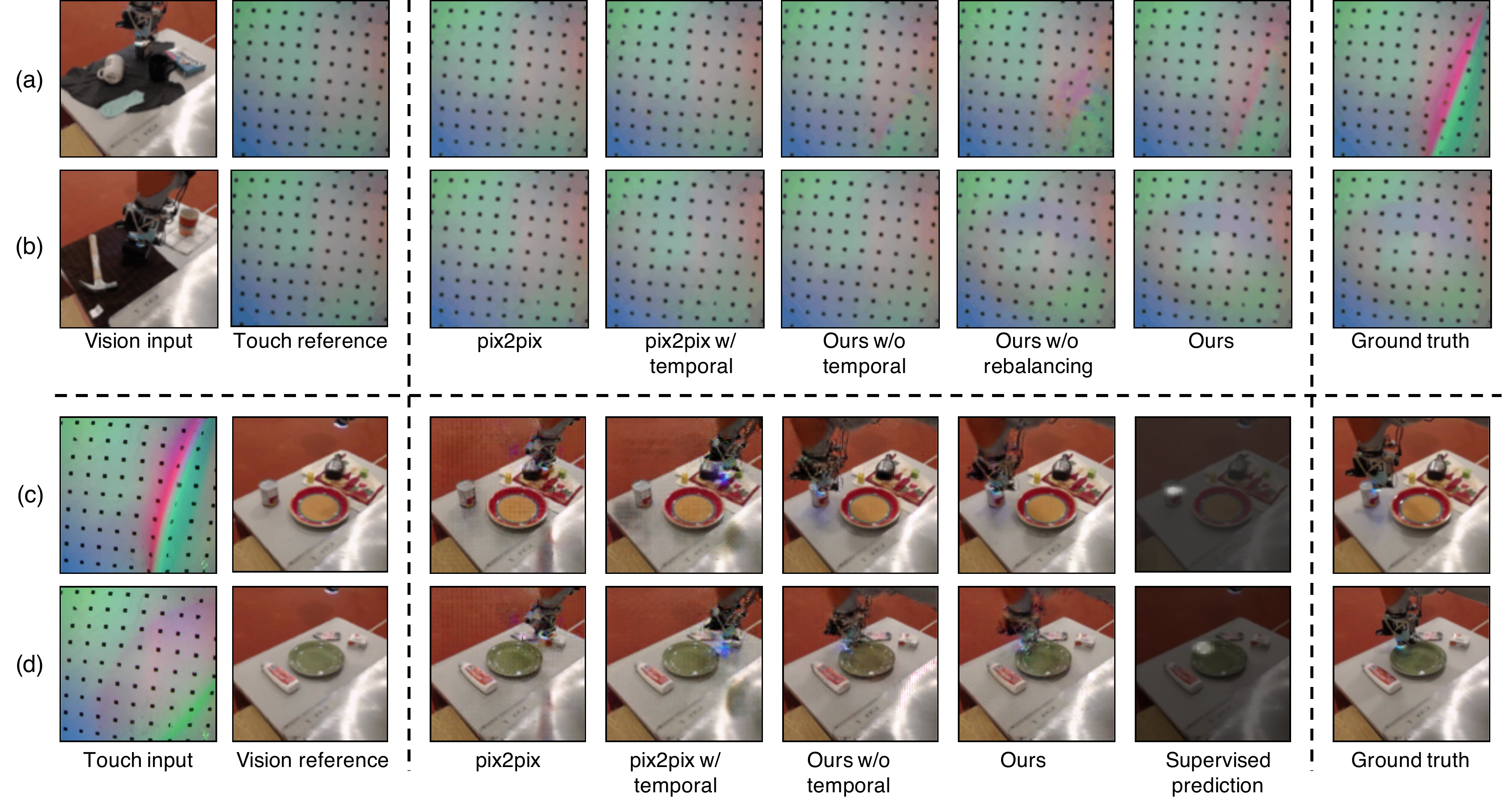}
\vspace{-15pt}
\caption{\small \lblfig{qualitative} {\bf Example cross-modal prediction results.} (a) and (b) show two examples of vision $\rightarrow$ touch prediction by our model and baselines. (c) and (d) show the touch $\rightarrow$ vision direction. In both cases, our results appear both realistic and visually similar to the ground truth target images. In (c) and (d), our model,  trained without ground truth position annotation, can accurately predict touch locations, comparable to a fully supervised prediction method. 
}
\lblfig{quali}
\vspace{-10pt}
\end{figure*}

\subsection{Implementation details}
\lblsec{method:details}
\vspace{10pt}
\myparagraph{Network architectures}
\lblsec{arch}
We use an encoder-decoder architecture for our generator. For the encoder, we use two ResNet-18 models~\cite{he2016deep} for encoding input images $\x$ and reference tactile and visual images $\rrr$ into $512$ dimensional latent vectors respectively. We concatenate two vectors from both encoders into one $1024$ dimensional vector and feed it to the decoder that contains $5$ standard strided-convolution layers. As the output result looks close to one of the reference images, we add several skip connections between the reference branch in the encoder and the decoder. For the discriminator, we use a standard ConvNet. Please find more details about our architectures in our supplement. 

\myparagraph{Training}
\lblsec{train}

We train the models with the Adam solver~\cite{kingma2014adam} with a learning rate of $0.0002$. We set $\lambda = 10$ for $\mathcal{L}_1$ loss.  We use LSGANs loss~\cite{mao2017least} rather than standard GANs~\cite{goodfellow2014generative} for more stable training, as shown in prior work~\cite{zhu2017unpaired,wang2017high}. We apply standard data augmentation techniques~\cite{krizhevsky2012imagenet} including random cropping and slightly perturbing the brightness, contrast, saturation, and hue of input images.

\section{Experiments}
\lblsec{experiments}
\vspace{-2mm}

We evaluate our method on cross-modal prediction tasks between vision and touch using the VisGel dataset. We report multiple metrics that evaluate different aspects of the predictions. For vision $\rightarrow$ touch prediction, we measure (1) perceptual realism using AMT: whether results look realistic, (2) the moment of contact: whether our model can predict if a GelSight sensor is in contact with the object, and (3) markers' deformation: whether our model can track the deformation of the membrane.  Regarding touch $\rightarrow$ vision direction, we evaluate our model using (1) visual realism via AMT and (2) the sense of touch: whether the predicted touch position shares a similar feel with the ground truth position. We also include the evaluations regarding full-reference metrics in the supplement. 
We feed reference images to all the baselines, as they are crucial for handling the scale discrepancy (\reffig{reference}). Please find our code, data, and more results on our \href{http://visgel.csail.mit.edu}{website}.

\subsection{Vision $\rightarrow$ Touch}
We first run the trained models to generate GelSight outputs frame by frame and then concatenate adjacent frames together into a video. Each video contains exactly one action with $64$ consecutive frames.

An ideal model should produce a perceptually realistic and temporal coherent output. Furthermore, when humans observe this kind of physical interaction, we can roughly infer the moment of contact and the force being applied to the touch; hence, we would also like to assess our model's understanding of the interaction. In particular, we evaluate whether our model can predict the moment of contact as well as the deformation of the markers grid. Our baselines include \pp~\cite{isola2017image} and different variants of our method. 
\myparagraph{Perceptual realism (AMT)}

\begin{table}
\centering
\scalebox{0.8} {
\begin{tabular}{lcc}
 & \textbf{Seen objects}  &  \textbf{Unseen objects} \\ 
\textbf{Method} &
\begin{tabular}{@{}c@{}}\% Turkers labeled \\ \emph{real}\end{tabular} &
\begin{tabular}{@{}c@{}}\% Turkers labeled \\ \emph{real}\end{tabular} \\
\hline
\pp~\cite{isola2017image} & 28.09 \% & 21.74\%  \\
\pp~\cite{isola2017image} w/ temporal & 35.02\%  & 27.70\%\\
Ours w/o temporal & 41.44\% & 31.60\% \\
Ours w/o rebalancing & 39.95\%  & 34.86\%  \\
{\bf Ours} & {\bf 46.63\% } &  {\bf 38.22\% } \\
\end{tabular} }
\vspace{2pt}
\caption {{\bf Vision2Touch AMT ``real vs fake" test.} Our method can synthesize more realistic tactile signals, compared to both \pp~\cite{isola2017image} and our baselines, both for seen and novel objects.}
\vspace{-15pt}
\lbltbl{AMT1}
\end{table}

Human subjective ratings have been shown to be a more meaningful metric for image synthesis tasks~\cite{zhang2016colorful,isola2017image} compared to metrics such as RMS or SSIM. We follow a similar study protocol as described in Zhang et al.~\cite{zhang2016colorful} and run a real vs. fake forced-choice test on  Amazon Mechanical Turk (AMT).   
In particular, we present our participants with the ground truth tactile videos and the predicted tactile results along with the vision inputs. We ask which tactile video corresponds better to the input vision signal. As most people may not be familiar with tactile data,  we first educate the participants with $5$ typical ground truth vision-touch video pairs and detailed instruction.
In total, we collect $8,000$ judgments for $1,250$ results.  \reftbl{AMT1} shows that our full method can outperform the baselines on both seen objects (different touches) and unseen objects.

\myparagraph{The moment of contact}
The deformation on the GelSight marker field indicates whether and when a GelSight sensor touches a surface. We evaluate our system by measuring how well it can predict the \textit{moment of contact} by comparing with the ground truth contact event. We track the GelSight markers and calculate the average $L_2$ distance for each marker. For each touch sequence, we denote the largest deformation distance as $d_\text{max}$ and the smallest deformation as $d_\text{min}$, we then set a cutoff threshold at $r\cdot(d_\text{max} - d_\text{min}) + d_\text{min}$, where $r$ is set to $0.6$ in our case. We mark the left most and right most cutoff time point as $t_l$ and $t_r$ individually. Similarly, we compute the ground truth cutoff time as $t_l^\text{gt}$ and $t_r^\text{gt}$; then the error of the \textit{moment of contact} for this sequence is determined as $|t_l - t_l^\text{gt}| + |t_r - t_r^\text{gt}|$.

\begin{figure}[t]
\centering
\includegraphics[width=.48\textwidth]{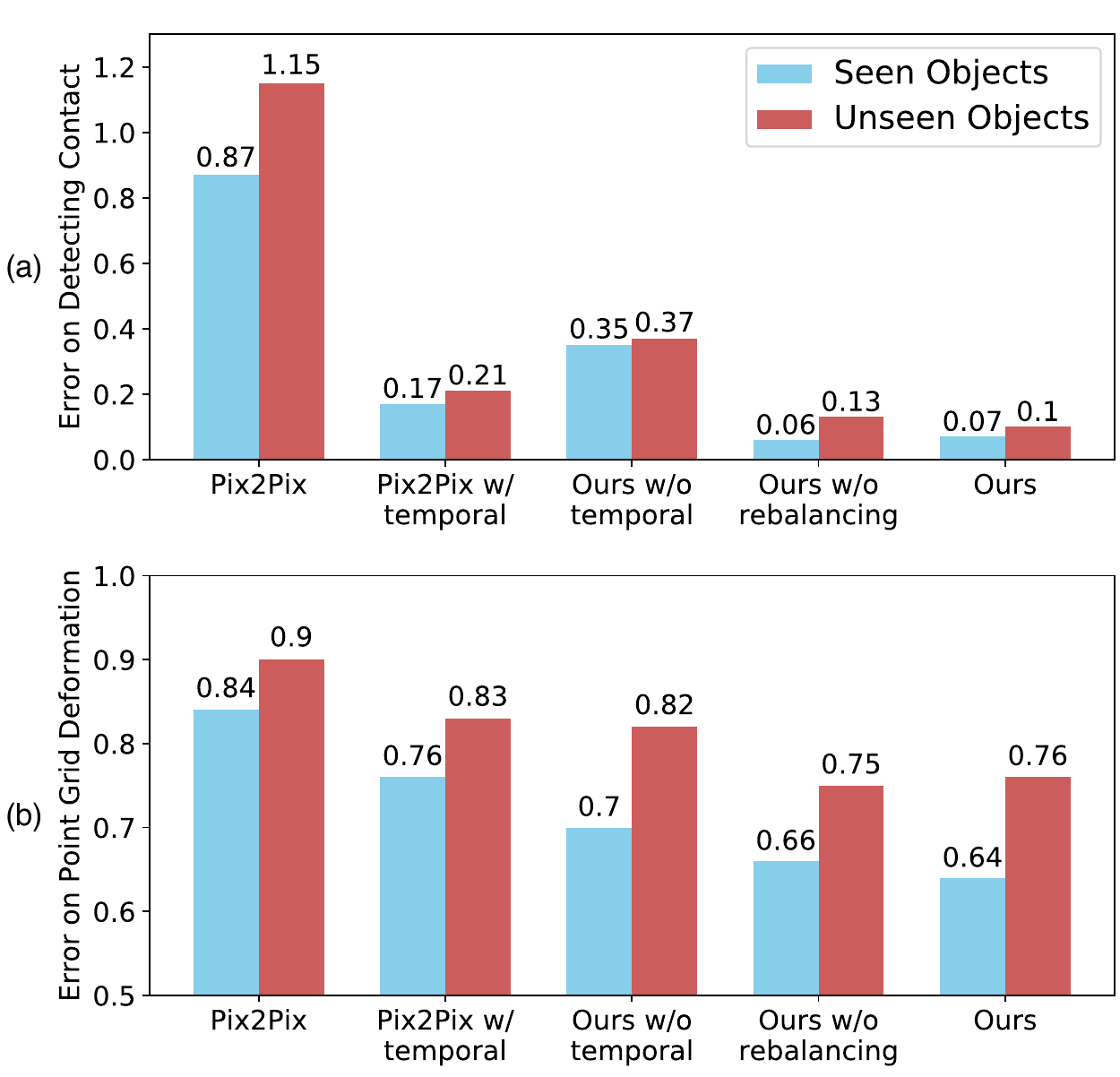}
\vspace{-10pt}
\caption{\small \label{fig:vision2touch_quantitative} {\bf Vision2Touch - quantitative results.} {\bf Top:} Errors on detecting the moment of contact. Our method generally performs the best. The use of temporal cues can significantly improve the performance or our model. {\bf Bottom:} Errors on the average markers' deformation. Our method still works best. }
\vspace{-10pt}
\end{figure}

\begin{figure}[!ht]
\centering
\includegraphics[width=.48\textwidth]{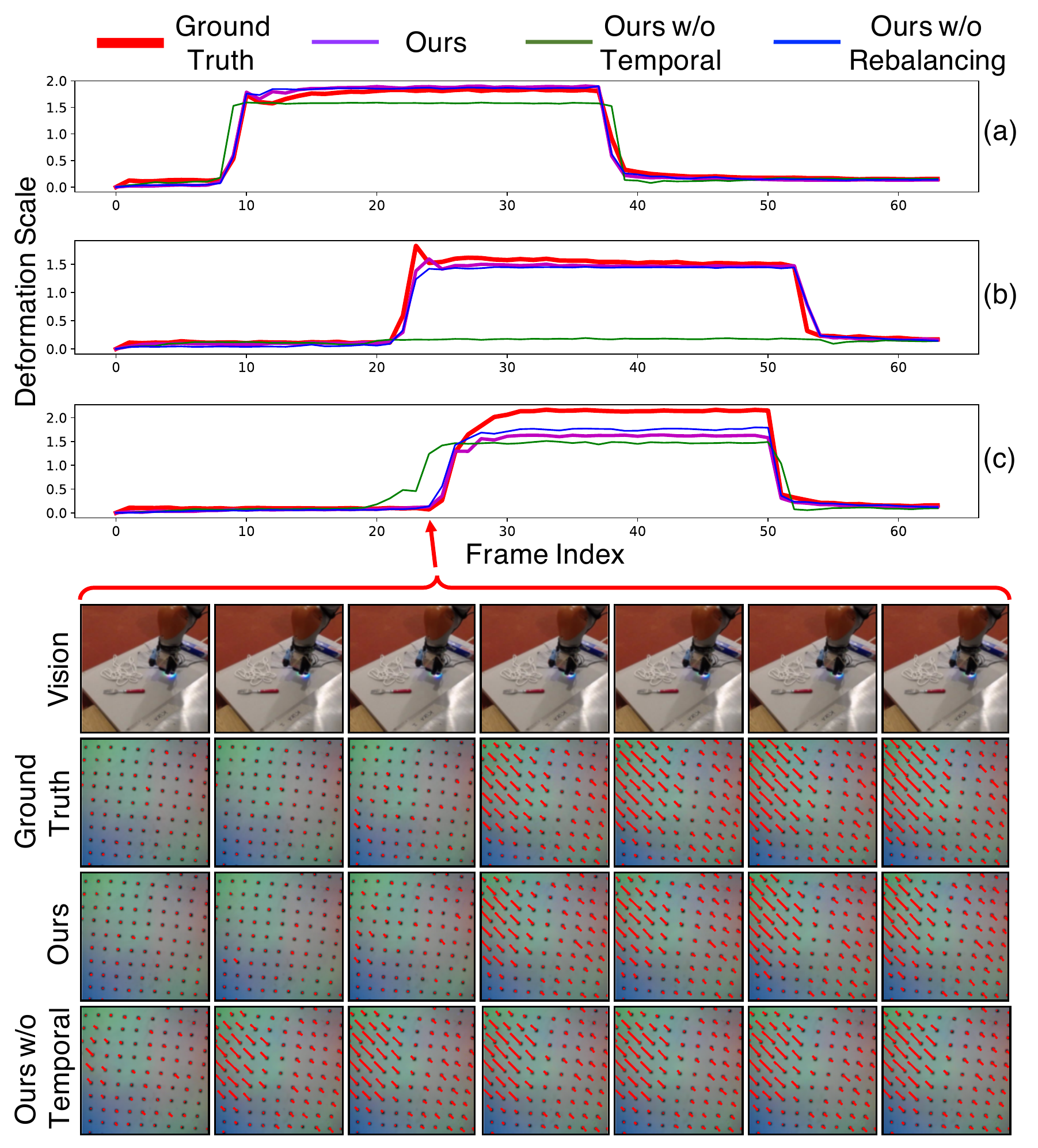}
\vspace{-15pt}
\caption{\small{\bf Vision2Touch - detecting the moment of contact.} We show the markers' deformation across time, determined by the average shift of all black markers. Higher deformation implies object contact with a larger force. {\bf Top:} Three typical cases, where (a) all methods can infer the moment of contact, (b) the method without temporal cues failed to capture the moment of contact, and (c) the method without temporal cues produces misaligned results. {\bf Bottom:} We show several  vision and touch frames from case (c). Our model with temporal cues can predict GelSight's deformation more accurately. The motion of the markers is magnified in red for better visualization.
}
\lblfig{motion_scale} 
\vspace{-10pt}
\end{figure}

As shown in \reffig{vision2touch_quantitative}a, the methods without temporal cues produce a large error due to temporal misalignment. \reffig{vision2touch_quantitative}a also shows that our model works better on seen objects than unseen objects, which coincides with the empirical observation that humans can better predict the touch outcomes if they have interacted with the object before.

We also show deformation curves for several torch sequences. \reffig{motion_scale}a illustrates a case where all the methods perform well in detecting the ground truth \textit{moment of contact}. \reffig{motion_scale}b shows an example where the model without temporal cues completely missed the contact event. \reffig{motion_scale}c shows another common situation, where the \textit{moment of contact} predicted by the method without temporal cues shifts from the ground truth. 
\reffig{motion_scale} shows several groundtruth frames and predicted results. As expected, a single-frame method fails to accurately predict the contact moment.

\myparagraph{Tracking markers' deformation}
The flow field of the GelSight markers characterizes the deformation of the membrane, which is useful for representing contact forces as well as detecting slippery~\cite{dong2017improved}. In this section, we assess our model's ability by comparing the predicted deformation with the ground truth deformation.
We calculate the average $L_2$ distance between each corresponding markers in the ground truth and the generated touch image.  \reffig{vision2touch_quantitative}b shows that the single-frame model performs the worst, as it misses important temporal cues, which makes it hard to infer information like force and sliding.

\begin{figure}[t]
\centering
\includegraphics[width=.45\textwidth]{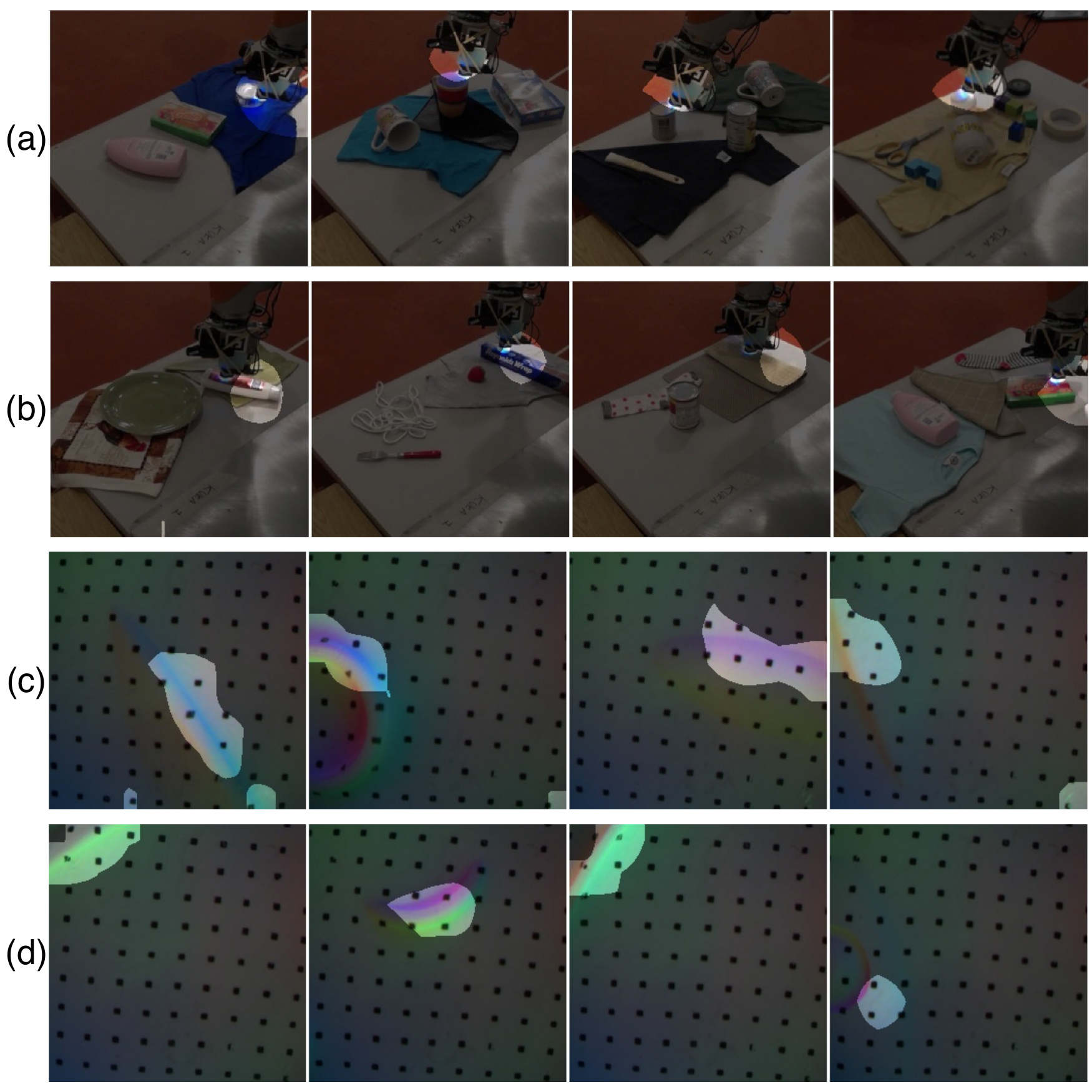}
\caption{\small {\bf Visualizing the learned representations using Zhou et al.~\cite{zhou2014object}}
(a) and (b) visualize two internal units of our vision $\rightarrow$ touch model. They both highlight the position of the GelSight, but focus on sharp edges and flat surfaces respectively. (c) and (d) visualize internal units of our touch $\rightarrow$ vision model. They focus on different geometric patterns.}
 \lblfig{vis} 
\vspace{-10pt}
\end{figure}

\myparagraph{Visualizing the learned representation}
We visualize the learned representation using a recent network interpretation method~\cite{zhou2014object}, which can highlight important image regions for final decisions (\reffig{vis}a and b). Many meaningful patterns emerge, such as arms hovering in the air or touching flat area and sharp edges. This result implies that our representation learns shared information across two modalities. Please see our \supp for more visualizations.

\subsection{Touch $\rightarrow$ Vision}
We can also go from touch to vision - by giving the model a reference visual image and tactile signal, can the model imagine where it is touching? 
It is impossible to locate the GelSight if the sensor is not in contact with anything; hence, we only include vision-touch pairs where the sensor touches the objects. An accurate model should predict a reasonable touch position from the geometric cues of the touch images.

\myparagraph{The Sense of Touch}
Different regions and objects can stimulate similar senses of touch. For example, our finger may feel the same if we touch various regions on a flat surface or along the same sharp edge. Therefore, given a tactile input, it is unrealistic to ask a model to predict the exact same touch location as the ground truth. As long as the model can predict a touch position that feels the same as the ground truth position, it is still a valid prediction. 
To quantify this, we show the predicted visual image as well as the ground truth visual image. Then we ask human participants whether these two touch locations feel similar or not. We report the average accuracy of each method over $400$ images ($200$ seen objects and  $200$ unseen objects). Table~\ref{tab:touch} shows the performance of different methods. Our full method can produce much more plausible touch positions.

We also compare our method with a baseline trained with explicit supervision provided by humans. Specifically, we hand-label the position of the GelSight on $1,000$ images, and train a Stacked Hourglass Network~\cite{newell2016stacked} to predict plausible touch positions. Qualitative and quantitative comparisons are shown in~\reffig{quali} and Table~\ref{tab:touch}. Our self-supervised method is comparable to its fully-supervised counterpart.

\begin{table}
\centering
\scalebox{0.8} {
\begin{tabular}{clcc}
 & & \textbf{Seen objects}  &  \textbf{Unseen objects} \\ 
 &\textbf{Method} & 
 \begin{tabular}{@{}c@{}}\% Turkers labeled \\ \emph{Feels Similar}\end{tabular}  & 
 \begin{tabular}{@{}c@{}}\% Turkers labeled \\ \emph{Feels Similar}\end{tabular} \\
\hline
\multirow{4}{*}{\rotatebox[origin=c]{90}{\parbox[c]{1.5cm}{\centering Self-Supervised}}}
 & \pp~\cite{isola2017image} & 44.52\% & 25.21\%  \\
 & \pp w/ temporal & 53.27\%  & 35.45\%\\
 & Ours w/o temporal & 81.31\% & 78.40\% \\
 & {\bf Ours} & {\bf 89.20\%} &  {\bf 83.44\%} \\
\hline
 & Supervised prediction & {\bf 90.37\%} & {\bf 85.29\%} \\
\end{tabular} }
\vspace{2pt}
\caption {{\bf Touch2Vision ``Feels Similar vs Feels Different" test}. Our self-supervised method significantly outperforms baselines. The accuracy is comparable to fully supervised prediction method trained with ground truth annotations.}
\vspace{-5pt}
\label{tab:touch}
\end{table}

\myparagraph{Perceptual Realism (AMT)}
Since it is difficult for humans participants to imagine a robotic manipulation scene given only tactile data, we evaluate the quality of results without showing the tactile input.
In particular, we show each image for $1$ second, and AMT participants are then given unlimited time to decide which one is fake. The first $10$ images of each HITs are used for practice and we give AMT participants the correct answer. The participants are then asked to finish the next $40$ trials.

In total, we collect $8,000$ judgments for $1,000$ results. Table~\ref{tab:AMT} shows the fooling rate of each method. We note that results from pix2pix~\cite{isola2017image} suffer from severe mode collapse and always produce identical images, although they look realistic according to AMT participants. See our website for a more detailed comparison. We also observe that the temporal cues do not always improve the quality of results for touch $\rightarrow$ vision direction as we only consider vision-touch pairs in which the sensor touches the objects. 

\begin{table}[t]
\centering
\scalebox{0.8} {
\begin{tabular}{lcc}
 & \textbf{Seen objects}  &  \textbf{Unseen objects} \\ 
\textbf{Method} & 
\begin{tabular}{@{}c@{}}\% Turkers labeled \\ \emph{real}\end{tabular} & 
\begin{tabular}{@{}c@{}}\% Turkers labeled \\ \emph{real}\end{tabular} \\
\hline
\pp~\cite{isola2017image} & 25.80\% &{\bf 26.13\%}  \\
\pp~\cite{isola2017image} w/ temporal & 23.61\%  &  19.67\%\\
Ours w/o temporal & {\bf 30.80\%} & 20.74\% \\
{\bf Ours} & 30.50\%  &   24.22\%  \\
\end{tabular} }
\vspace{2pt}
\caption {{\bf Touch2Vision AMT ``real vs fake" test}. Although pix2pix achieves the highest score for unseen objects, it always produces identical images due to mode collapse. \reffig{quali} shows a typical collapsed mode, where pix2pix always places the arm at the top-right corner of the image. More qualitative results can be found in our supplementary material.}
\vspace{-10pt}
\label{tab:AMT}
\end{table}

\myparagraph{Visualizing the learned representation}
The visualization of the learned representations (\reffig{vis}c and d) show two units that focus on different geometric cues. Please see our \supp for more examples.
\section{Discussion}
\lblsec{conclusion}
\vspace{-2mm}

In this work, we have proposed to draw connections between vision and touch with conditional adversarial networks. Humans heavily rely on both sensory modalities when interacting with the world. Our model can produce promising cross-modal prediction results for both known objects and unseen objects. In the future, vision-touch cross-modal connection may help downstream vision and robotics applications, such as object recognition and grasping in a low-light environment, and physical scene understanding.

\myparagraph{Acknowledgement}
We thank Shaoxiong Wang, Wenzhen Yuan, and Siyuan Dong for their insightful advice and technical support. This work was supported by Draper Laboratory Incorporated (Award No. SC001-0000001002) and NASA-Johnson Space Center (Award No. NNX16AC49A).

{\small
\bibliographystyle{ieee_fullname}
\bibliography{egbib}
}

\end{document}